%% file: ijcai21-multiauthor.tex
\documentclass{article}
\usepackage{ijcai21}

\usepackage{times}

\usepackage{soul}
\usepackage{url}
\usepackage[hidelinks]{hyperref}
\usepackage[utf8]{inputenc}
\usepackage[small]{caption}
\usepackage{graphicx}
\usepackage{amsmath}
\usepackage{booktabs}
\usepackage{multirow}

\usepackage{hyperref}
\usepackage{xurl}

\title{Can Self Reported Symptoms Predict Daily COVID-19 Cases?}

\author{ Parth Patwa
$^1$\footnote{Contact Author}\and
Viswanatha Reddy$^1$\and
Rohan Sukumaran$^{1}$\and
Sethuraman TV$^{1}$\and Eptehal Nashnoush$^{1}$\and Sheshank Shankar$^{1}$\and Rishemjit Kaur$^{1,2}$ \and Abhishek Singh$^{1,3}$\And Ramesh Raskar$^{1,3}$ 
 \\
\affiliations
$^1$PathCheck Foundation\\
$^2$ CSIR-Central Scientific Instruments Organisation\\
$^3$MIT Media Lab\\

\emails
parth.patwa@pathcheck.org
}

\begin{document}

\maketitle


\begin{abstract}

The COVID-19 pandemic has impacted lives and economies across the globe, leading to many deaths. While vaccination is an important intervention, its roll-out is slow and unequal across the globe. Therefore, extensive testing still remains one of the key methods to monitor and contain the virus. Testing on a large scale is expensive and arduous. Hence, we need alternate methods to estimate the number of cases. Online surveys have been shown to be an effective method for data collection amidst the pandemic. In this work, we develop machine learning models to estimate the prevalence of COVID-19 using self-reported symptoms. Our best model predicts the daily cases with a mean absolute error (MAE) of $226.30$ (normalized MAE of 27.09\%) per state, which demonstrates the possibility of predicting the actual number of confirmed cases by utilizing self-reported symptoms. The models are developed at two levels of data granularity - local models, which are trained at the state level, and a single global model which is trained on the combined data aggregated across all states. Our results indicate a lower error on the local models as opposed to the global model. In addition, we also show that the most important symptoms (features) vary considerably from state to state. This work demonstrates that the models developed on crowd-sourced data, curated via online platforms, can complement the existing epidemiological surveillance infrastructure in a cost-effective manner. The code is publicly available at \url{https://github.com/parthpatwa/Can-Self-Reported-Symptoms-Predict-Daily-COVID-19-Cases}.
\end{abstract}

\input{introduction}
 \input{related_work}

 \input{data}
\input{methodology}
\input{results}

\input{conclusion}
\input{acknowledgement}

\bibliography{references}
\bibliographystyle{named}
\newpage

 \input{appendix}

\end{document}

%% file: introduction.tex
\section{Introduction}\label{Introduction} 
\frenchspacing

The COVID-19 pandemic has created turmoil in the lives of millions of people across the globe. Even with the advent of vaccines, their distribution remains a challenge \cite{bae2020challenges,samalvacc2021,mills2021challenges}. We need interventions, policies, and solutions that go beyond pharmaceutical interventions (vaccines, therapeutics, etc). Testing and identifying the number of COVID-19 cases is important for us to identify how well we are doing against the virus. Although important, even with rapid tests and at-home testing facilities, we are testing less than 5.61 per 1000 of the population in the US \cite{ourworldindata,morales2021covid,gandhi2020clinical}. Furthermore, getting test results takes time in hours, which can lead to delays in getting medical treatment. This calls for a non-invasive, and scalable method for estimating the cases, to complement traditional testing infrastructure. In this paper, we harness the power of deep learning and crowd-sourced symptoms data for the prediction of COVID 19 cases on a daily basis. For this purpose, we use the data collected by CMU via Facebook surveys \cite{CMUDataset}. This data of self-reported symptoms is available at the state level for the US. We integrate this self-reported symptoms data with the actual COVID-19 cases reported by WHO. Here, we answer an important question \textit{``Can we predict daily COVID-19 cases at the granularity of a state, using self-reported symptoms from a fraction of the population?''}.


We predict the actual daily case count, using self-reported symptoms data that is aggregated at the population level (to obfuscate personal information), by training machine learning (ML) and deep learning (DL) algorithms. We train models at two levels of data granularity - global and local. At the global level, we combine the data from all the states into a single data set and train a single model on it, whereas at the local level, we train a separate model for each state. At the global level, the best model has a mean absolute error (MAE) of 368.26 and normalized MAE (nMAE) of 44.08\%. Further, when we take an ensemble of all the local models, the \textbf{MAE reduces to 226.30 per state} (nMAE = 27.09\%). This indicates that the model is off by 226 cases per state when predicting daily cases. The success of our experiments highlights the fact that our model can serve as a low resource way of estimating COVID-19 cases. We observe that the top features contributing to predictions vary from state to state, alluding to the benefits of local models. Our model may be used to detect hot spots and can act as a non-obtrusive, economical, and effective screening method in a low resource environment.





%% file: related_work.tex
\section{Related Work}
\frenchspacing
Recently, there has been significant traction in research related to COVID-19 both in terms of clinical and digital innovations, where the scientific community has focused on this disease with near-unprecedented intensity. Along the clinical direction, there were several efforts like vaccine development \cite{KAUR2020198114}, re-purposing known clinically-tested drugs, and virtual screening for possible targets using protein structure data \cite{10.3389/frai.2020.00065,PPR:PPR112521}, understanding efficacy of testing \cite{Jarrom2020.08.10.20171777}, etc. Efforts along clinical and other medical directions are instrumental but can be time-consuming. The drastic increase in cases has challenged the medical infrastructure worldwide in various aspects, including a sharp rise in demand for hospital beds, shortage of medical equipment and personnel \cite{sen2021closer}. At the same time, testing methods are also facing an acute shortage in developing countries. Thereby causing a delay in getting test results leading to increased infection rates and delays in critical preventive measures. Thus judicial usage of health care resources like testing, vaccines is crucial.

To complement the medical research with computational solutions, efforts have been made on predictive modelling of the disease spread, simulations on vaccine distribution strategies, etc \cite{Romero-Brufaun1087}. Many efforts were along the directions of understanding the severity, spread, and unique characteristics of the COVID-19 infection, across a broad range of clinical, imaging, and population-level datasets \cite{Gostic,Liang,Menni,Shi,shankar2020proximity}. For instance, various studies have tried to understand the progression of the virus, future hot-spot, estimating the number of cases/deaths/hospitalization, etc. using exposure notification \cite{DBLP:journals/corr/abs-2006-08543}, epidemiological modeling \cite{Romero-Brufaun1087}. Studies have also tried mathematical modeling to understand the outbreak under different situations for different demographics \cite{menni2020real,saad2020immune,wilder2020tracking}. Apart from these, several machine learning models were developed to forecast the COVID-19 cases in regions like India, Egypt \cite{FAROOQ2021587,AMAR2020622}, etc.  



 

Effective screening enables quick and efficient diagnosis of COVID-19 and can mitigate the burden on public healthcare systems. Prediction models which combine several features (symptoms, testing, mobility, etc.) to estimate the risk of infection have been developed to assisting medical staff worldwide in triaging patients. Some of these models use laboratory tests \cite{Feng2020.03.19.20039099,mei2020artificial}, clinical symptoms \cite{tostmann2020strong}, and integration of both \cite{punn2020covid}. However, most previous models were based on data from hospitalized patients and thus are not effective in screening for COVID-19 in the general population. 

Hence, the development of non-obtrusive system disentangled from the health care infrastructure becomes imperative to accelerate the efforts against COVID-19. \cite{sukumaran2020covid19} used self-reported symptoms to predict outbreaks and is the closest to our work. However, unlike their work, we predict \textit{actual daily cases} instead of \textit{self-reported daily cases}.


%% file: data.tex
\section{Data}\label{Data}

\textbf{CMU dataset} - We use the symptoms survey data from CMU \cite{CMUDataset}. This was a survey collected across ~70,000 on a daily basis. It consists of multiple questions along the directions of symptoms (cough, fever, etc), behavioral patterns (working outside, avoid contact, etc), medical conditions (cancer, heart disease, etc), and more. They aggregate the data and provide it in the form of percentage of respondents who reported having that symptom/behavior. The data is available at the county and state level with a total of 104 features (as of October 2020), including weighted (adjusted for sampling bias), unweighted signals, demographic columns (age, gender, etc). We use the state level data from Apr. 4, '20 to Sep. 11, '20. 

\textbf{Daily Cases} - NY Times \cite{nytimes} reports the cumulative number of cases in a state on a given date, as provided by WHO. From this data, we compute and use the daily new cases in a state.





%% file: methodology.tex
\section{Methodology and Experiments}\label{Methodology}

We predict the daily cases of COVID-19 in the US states \cite{nytimes} using data from 6 April 2020 to 11 September 2020.    

\textbf{Input Features} - We use features provided in the CMU  \cite{CMUDataset} dataset.  We follow a feature selection and ranking process similar to \cite{sukumaran2020covid19}. Further, we prune un-weighted and other signals (age, gender, derived features, etc) which leaves us with 35 features. We further rank these 35 features according to their f\_regression \cite{Freg} scores against the target variable, and then input them to the models. As the demographic level (age, gender) split of actual daily cases is not available, we drop such data points from the CMU data. Thereby, only using data points that aggregate all demographics (gender and age).

\textbf{Data Granularity levels} - We train on 2 levels: \begin{itemize}
    \item Global - The data of all the states is combined and a single model is trained.    
    \item Local - A separate model is trained for every state. For comparison with the global level model, we ensemble the predictions from all the local models - for each state we take the test prediction from the respective local models. Thereby getting the local predictions from the respective local model. Later, we compute the error metrics for the entire test set. 
\end{itemize}

\textbf{Train-Test Split } - We split the entire data into train/test set on the basis of dates. The train data has the initial 80\% of the dates whereas the test data has the last 20\% of the dates. This ensures that the model has not seen future dates while training. It also ensures that the train and test data of every state is the same in the local and global level model.  



\textbf{Algorithms} - We experiment with five ML baselines - Linear Regression (LR), Decision Tree (DT), Multi Layer Perceptron (MLP), Gradient Boost (GDBT), and XGBoost (XGB). We also implement 2 DL models: \begin{itemize}
    \item CNN - An architecture with seven convolutional layers followed by dense layers. 
    
    \item \textbf{1d Resnet} - Inspired by the wide success of ResNets \cite{He2015} in various fields we have developed a model similar to ResNet-18 for one-dimensional data. Our proposed architecture comprises of 3 blocks.
    At  the  end  of  the  network, a Global Average Pooling (GAP) layer is used followed by fully connected layers. The dense layers comprises of 256, 128, and 1 neuron respectively. Between each fully connected layer, a dropout layer is employed with a probability of 0.5 (p=0.5).  
\end{itemize}   

\textbf{Error Metric} - We use two error metrics to evaluate and compare our models: \begin{itemize}
    \item Mean Absolute Error (MAE) - The average over all data points of the absolute value of the difference between the predicted value and the actual value. \\
            MAE = $\frac{1}{n}\sum_{i=1}^{n}|p_i - t_i|$
     
       Where n is the total data instances, $p_i$ is the predicted value and $t_i$ is the actual value (ground truth).

    \item Normalized Mean Absolute Error (nMAE) - Normalized error is calculated to capture the variation in the number of daily cases across different dates and states. \\
            nMAE = $100* \frac{\sum_{i=1}^{n}|p_i - t_i|}{\sum_{i=1}^{n}t_i} \%$

\end{itemize}
To show statistical significance, we also calculate the 95\% confidence interval (CI) over 20 runs (the random seed is changed every time) for our ML based global level models. 

\textbf{Implementation} - We use Scikit-learn \cite{scikit-learn}, Keras \cite{chollet2015keras}, and xgboost library \cite{Chen2016xgb} for our implementation. The ML models are trained on an Intel i7 8th generation CPU, while DL models are trained on Google Colab. The code is publicly available at \url{https://github.com/parthpatwa/Can-Self-Reported-Symptoms-Predict-Daily-COVID-19-Cases}. 

%% file: results.tex
\section{Results and Analysis}\label{Results}
\frenchspacing
\begin{table*}[ht!]
\centering
\begin{tabular}{|l|l|l|l|l|}
\hline

Algorithm & nMAE & MAE & nMAE CI & MAE CI \\ \hline
LR        &    98.84\%  &  826.05   &   (98.84, 98.84)      &  (826.05, 826.05)      \\ \hline
DT        &  72.87\%    &   609.02  &    (72.48, 73.27)     &   (605.76, 612.28)     \\ \hline
MLP       &  66.30\%    & 554.06    &    (64.90, 67.70)     &  (542.37, 565.74)      \\ \hline
GDBT      &   60.15\%   &   502.73  &  (60.13, 60.18)       &  (502.54, 502.91)      \\ \hline
XGB       &  54.10\%    &  452.12   &     (53.92, 54.28)    &   (450.62, 453.62)     \\ \hline
CNN       &    54.92\% &   459.09 &   -      &  -      \\ \hline
\textbf{1d Resnet} &   \textbf{44.08\%}    & \textbf{368.26}      &      -  &  -      \\ \hline
\end{tabular}
    \caption{Test data results for prediction of daily cases per state by various models trained on the global level. The 95\% confidence interval (CI) is calculated on 20 runs (random seed changed every time). The models use all the 35 features. }
    \label{table:symptoms}
\end{table*}

 \begin{table*}[ht]
\begin{tabular}{|l|l|}
\hline
Model  & Top 5 features                                                                                                     \\ \hline
ca     & cmnty cli, avoid contact all or most time, hh fever, hh sore throat, hh cough                                      \\ \hline
tx     & cmnty cli, anosmia ageusia, avoid contact all or most time, worked outside home, high blood pressure     \\ \hline
fl     & cmnty cli, anosmia ageusia, avoid contact all or most time, worked outside home, persistent pain pressure in chest \\ \hline
ak     & cmnty cli, worked outside home, avoid contact all or most time, self runny nose, none of above                     \\ \hline
vt     & none of above, hh cough, cancer, nasal congestion, avoid contact all or most time                                  \\ \hline
wy     & avoid contact all or most time, self shortness of breath, cmnty cli, other, tiredness or exhaustion                \\ \hline
Global & cmnty cli, anosmia ageusia, hh fever, hh cli, hh sore throat                                                       \\ \hline
\end{tabular}
\caption{Top 5 features for local and global level XGB models. Notice the lack of uniformity in top features across the states. cmnty - Community, cli - Covid Like Illness, hh - household. }
\label{tab:top_featues}
\end{table*}

\begin{table}[]
\centering
\begin{tabular}{|l|l|l|}
\hline
Algorithm & nMAE & MAE  \\ \hline
LR        & 45.65\%     &  381.51                   \\ \hline
DT        &   44.24 \%  &   369.75                    \\ \hline
MLP       & 36.70 \%    & 306.70                 \\ \hline
GDBT      &  36.68 \%    &   306.55                    \\ \hline
XGB       &   34.42\%   &   287.63                 \\ \hline
CNN       &   101.00\%   &  849.60        \\ \hline
\textbf{1d Resnet}  & \textbf{27.09\%}  &     \textbf{226.30}         \\ \hline
\end{tabular}
    \caption{Test results for prediction of daily cases per state by models trained on the local level. A local model is trained for each state. The results in this table are of the ensemble of all local level models. }
    \label{table:granular}
\end{table}

\begin{figure}[]
    \centering
    \includegraphics[width = 0.6\linewidth]{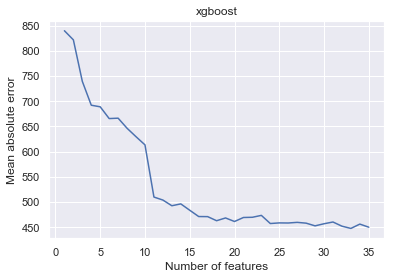}
  \caption{MAE vs the number of features used for XGB. In general, the error decreases with increase in the number of features. The rate of decrease in error decreases with increase in the number of features.  }%
  \label{fig:combined}
\end{figure}

\begin{table}[]
\centering
\begin{tabular}{|l|l|l|l|l|}
\hline
State   & \multicolumn{2}{l|}{MAE} & \multicolumn{2}{l|}{nMAE (\%)} \\ \hline
      & local      & global      & local       & global      \\ \hline 
     tx &1231.87 &1831.06 &23.09 &34.32 \\ \hline
ca      &  1338.76          &      1601.92       &    18.82         &   22.52          \\ \hline
fl      &   1275.58         &    2025.70         &  32.63           & 51.81            \\ \hline
wy      &  15.32          &   23.38          &     43.65       &   66.60          \\ \hline
me      &   7.79         &   31.70          &  34.83           & 141.75            \\ \hline
vt      &  3.48          &  12.87           &     53.59        &  197.96           \\ \hline
\textbf{entire} &  \textbf{226.30}          &   368.26         &   \textbf{27.09}          &  44.08           \\ \hline
\end{tabular}
\caption{MAE and nMAE of 1d resnet model trained on the local and global level for few states and the entire USA, on the test data. For all the states, please refer to Section \ref{Appendix} (Appendix).}
\label{tab:local_vs_global_main}
\end{table}

\begin{figure}[]
    \centering
    \includegraphics[width=\linewidth]{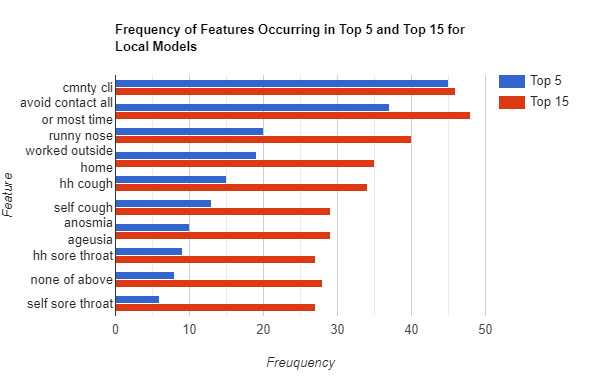}
  \caption{Some features with the frequency of their occurrence in the top 5 and top 15 important features in local models. Most features occur in the top 5 or 15 for only a few states, indicating high variance in top features across states. Data for all the features is reported in section \ref{Appendix} (appendix).   }%
  \label{fig:feature_count}
\end{figure}


Table \ref{table:symptoms} shows the result of daily cases prediction by different models trained on the global level. Our 1D Resnet performs the best with 382.40 MAE and 45.77\% nMAE per state. CNN performs worse than 1d resnet. Among ML models, XGB (MAE = 502.73 , nMAE = 60.15\%) performs the best, while  LR has the worst performance. 

Figure \ref{fig:combined} shows the MAE vs the number of features used for XGB. It can be observed that with the increase in number of features, MAE decreases. However, MAE plateaus after incorporating approx. 15 features. Consequently, we can even use fewer (around 15) features to predict daily cases without considerably increasing the error. This observation is similar to that of \cite{sukumaran2020covid19} and may aid in reducing the number of questions of the surveys. 

Table \ref{table:granular} shows the MAE of ensemble of models trained on the local level (one model per state). We see that 1d resnet performs the best and gives an \textbf{MAE of 226.30 per state across the USA} (nMAE = 27.09\%). LR performs the worst.

We can see from Table \ref{table:symptoms} and Table \ref{table:granular} that for all algorithms except CNN, the ensemble of local models achieves substantially lower error than a single global model. For example, the MAE of XGB trained on global level is 452.12 (nMAE =54.10\%) whereas MAE of XGB models trained on local level is 287.63 (nMAE = 34.43\%). The poor results of local CNN models could be due to lack of data per state or overfitting. CNN fails to capture complex relations that generalize well. The superiority of local models over the global model is further observed in Table \ref{tab:local_vs_global_main} which shows the state-wise MAE by 1d Resnet trained on global and local level for a few states. \textbf{Local level models outperform the the global level model in 49 states out of 50}. These observations motivate us to analyze the state-wise top contributing features.  

Figure \ref{fig:feature_count} shows the number of times a feature was in the top 5 or top 15 important features across the local models, based on the feature importance given by the XGB models. We see that \textit{cmnty cli} (covid like illness in community), and \textit{avoid contact all or most time} are important features for most of the local level models. However, the frequency of the remaining features being in the top 5 or top 15 features is low. 32 of the 35 features are present in top 5 for at least 1 state and all the features are in the top 15 for at least 9 states. This shows that the top important features vary considerably across the states. Table \ref{tab:top_featues} shows some states having different set of top 5 features. As the feature importance (or ranking) varies considerably across the states, local models perform better than the global model. This also means that COVID-19 affects every state differently. Further, we see that the 1st (\textit{cmnty cli}), 2nd \textit{avoid contact all or most time}), and 4th (\textit{worked outside home} most occurring top features are related to social distancing, which highlights the contagious nature of COVID-19.


%% file: conclusion.tex
\section{Conclusion and Future Work}
\frenchspacing
Forecasting the epidemic and its spread is an important tool in pandemic response. In this work, we assess the possibility of building an outbreak prediction system using crowd-sourced symptoms data. Our experiments demonstrate that self-reported symptoms can predict actual cases with low error. Furthermore, a small number of features (symptoms) are sufficient to predict the total number of cases with reasonable accuracy. The analysis suggests that learning models at a state level improves the prediction performance possibly due to the top features vary across different states. In other words, COVID-19 affects every state differently. This information can be used to create state-specific and shorter surveys. Consequently, with the increasing social media and internet penetration, this method can be scaled to complement physical testing facilities to estimate the number of cases, especially in low resource areas. Future directions worth exploring include, improving the prediction capability of the model by incorporating meta learning and transfer learning to improve performance for states with relatively lower number of samples, and to forecast cases as a time series. As the data collected here is health data, privacy-preserving machine learning could improve the adoption of such systems.


%% file: acknowledgement.tex
\section{Acknowledgement}
We thank Ishaan Singh for help in drafting this paper.

%% file: appendix.tex
\section{Appendix} \label{Appendix}
\begin{table}[]
\begin{tabular}{|l|l|l|l|l|}
\hline
\multirow{2}{*}{state} & \multicolumn{2}{l|}{MAE} & \multicolumn{2}{l|}{nMAE (\%)} \\ \cline{2-5} 
 & Local      & Global      & Local       & Global      \\ \hline
ak             & 20.57         & 34.487    & 27.85 & 46.70     \\ \hline
al             & 407.77         & 373.965   & 37.15 & 34.07      \\ \hline
ar             & 157.33           & 316.5366   & 29.30 & 58.96        \\ \hline
az             & 226.52         & 464.88    & 31.62 & 64.9       \\ \hline
ca             & 1338.76         & 1601.93   & 18.82 & 22.52      \\ \hline
co             & 65.25           & 94.07    & 19.49 & 28.10      \\ \hline
dc             & 18.56          & 23.43     & 34.61 & 43.69     \\ \hline
de             & 50.81         & 80.96    & 55.77 & 88.87      \\ \hline
fl             & 1275.58         & 2025.70    & 32.62 & 51.81       \\ \hline
ga             & 510.90         & 919.80    & 22.36 & 40.26     \\ \hline
hi             & 82.98         & 357.41     & 37.40 & 161.11     \\ \hline
ia             & 277.22         & 426.23    & 37.63 & 57.86       \\ \hline
id             & 96.08         & 146.48     & 30.18 & 45.47     \\ \hline
il             & 381.80         & 1000.20   & 19.30 & 50.57        \\ \hline
in             & 183.15         & 291.35    & 20.46 & 32.55     \\ \hline
ks             & 425.32         & 447.91     & 76.25 & 80.30     \\ \hline
ky             & 214.76         & 305.70      & 31.33 & 44.60    \\ \hline
la             & 409.52         & 495.45    & 53.85 & 65.15       \\ \hline
ma             & 366.00          & 425.01      & 571.28 & 663.38     \\ \hline
md             & 149.23         & 203.96      & 25.33 & 34.63    \\ \hline
me             & 7.79         &  31.70     & 34.83 & 141.74     \\ \hline
mi             & 208.71          & 323.62    & 28.26 & 43.83      \\ \hline
mn             & 149.04         & 319.29    & 21.57 & 46.22     \\ \hline
mo             & 210.50         & 379.88   & 16.99 & 30.67       \\ \hline
ms             & 181.35         & 234.99    & 26.92 & 34.88       \\ \hline
mt             & 34.42         & 117.85    & 29.86 & 102.27      \\ \hline
nc             & 331.10         & 526.08    & 23.15 & 36.78      \\ \hline
nd             & 75.68         & 128.36     & 36.17 & 61.35     \\ \hline
ne             & 90.33          & 177.09    & 33.68 & 66.03      \\ \hline
nh             & 7.81           & 12.63    & 35.36 & 57.13      \\ \hline
nj             & 123.61         & 164.06    & 37.74 & 50.09      \\ \hline
nm             & 49.90           & 195.38  & 38.72 & 148.96        \\ \hline
nv             & 168.63         & 502.96   & 31.84 & 94.97        \\ \hline
ny             & 145.12           & 169.44   & 22.05 & 25.75       \\ \hline
oh             & 189.54          & 282.73    & 18.25 & 27.23      \\ \hline
ok             & 159.91         & 341.58    & 22.32 & 47.68      \\ \hline
or             & 61.49         & 134.40   & 26.04 & 56.92       \\ \hline
pa             & 141.93         & 246.73    & 19.46 & 33.83      \\ \hline
ri             & 68.91         & 89.77    & 72.43 & 94.36      \\ \hline
sc             & 240.80         & 372.59    & 28.35 & 43.88     \\ \hline
sd             & 81.35         & 97.73    & 42.08 & 50.56        \\ \hline
tn             & 418.23          & 569.35    & 28.91 & 39.36      \\ \hline
tx             & 1231.87         & 1831.06   & 23.09 & 34.32       \\ \hline
ut             & 63.38          & 166.70    & 16.83 &    44.28   \\ \hline
va             & 188.00          & 476.48 & 18.47 & 46.81    \\ \hline
vt             & 3.48           & 12.87   & 53.59 & 197.96       \\ \hline
wa             & 135.77          & 209.61 & 25.77 & 39.80        \\ \hline
wi             & 151.44           & 385.41  & 19.84 & 50.50       \\ \hline
wv             & 41.85          & 104.18    & 31.64 & 78.78     \\ \hline
wy             & 15.32          & 23.38     & 43.65 & 66.60      \\ \hline
\textbf{Entire} & \textbf{226.301} & 368.26 & \textbf{27.09} & 44.08\\ \hline
\end{tabular}
\label{tab: appendix_global_local}
\caption{MAE and nMAE of local level 1d resnet vs global level 1d resnet for all states. Local models outperform the global model in 49 out of 50 states. }
\end{table}

\begin{table}[]
\begin{tabular}{|l|l|l|}
\hline
\textbf{Feature}                           & \textbf{Top 5} & \textbf{Top 15}\\ \hline
cmnty cli                         & 45    & 46     \\ \hline
avoid contact all or most time    & 37    & 48     \\ \hline
runny nose                        & 20    & 40     \\ \hline
worked outside home               & 19    & 35     \\ \hline
hh cough                          & 15    & 34     \\ \hline
self cough                        & 13    & 29     \\ \hline
anosmia ageusia                   & 10    & 29     \\ \hline
hh sore throat                    & 9     & 27     \\ \hline
none of above                     & 8     & 28     \\ \hline
self sore throat                  & 6     & 27     \\ \hline
multiple symptoms                 & 6     & 18     \\ \hline
nasal congestion                  & 5     & 28     \\ \hline
other                             & 5     & 23     \\ \hline
high blood pressure               & 5     & 16     \\ \hline
hh shortness of breath            & 5     & 18     \\ \hline
hh difficulty breathing           & 5     & 15     \\ \hline
hh cli                            & 5     & 18     \\ \hline
self difficulty breathing         & 4     & 18     \\ \hline
heart disease                     & 4     & 15     \\ \hline
persistent pain pressure in chest & 3     & 13     \\ \hline
muscle joint aches                & 3     & 13     \\ \hline
hh fever                          & 3     & 20     \\ \hline
self shortness of breath          & 2     & 20     \\ \hline
multiple medical conditions       & 2     & 15     \\ \hline
kidney disease                    & 2     & 10     \\ \hline
diarrhea                          & 2     & 22     \\ \hline
chronic lung disease              & 2     & 20     \\ \hline
tiredness or exhaustion           & 1     & 16     \\ \hline
self fever                        & 1     & 11     \\ \hline
no above medical conditions       & 1     & 12     \\ \hline
cancer                            & 1     & 12     \\ \hline
asthma                            & 1     & 9      \\ \hline
nausea vomiting                   & 0     & 14     \\ \hline
diabetes                          & 0     & 16     \\ \hline
autoimmune disorder               & 0     & 15     \\ \hline
\end{tabular}
\label{tab:appendix_feature}
\caption {All features and how often they occur in top 5 and top 15 features in local XGB models. Here, we can see that cmnty cli was among the top features in 45 out of 50 models and was among the top 15 features in 46 out of 50 models. cmnty - communty, cli - Covid Like Illness,  anosmia - loss of smell, ageusia - loss of taste, hh - Household, self - about the person taking the survey (individual level), none of above - none of the mentioned symptoms, other - other symptoms. }
\end{table}